\documentclass[nohyperref]{article}

\usepackage{microtype}
\usepackage{graphicx}
\usepackage{subcaption}
\usepackage{booktabs} 

\usepackage{hyperref}
\usepackage{etoolbox}

\usepackage[accepted]{arxiv_2022}

\usepackage{amsmath}
\usepackage{amssymb}
\usepackage{mathtools}
\usepackage{amsthm}

\usepackage[capitalize,noabbrev]{cleveref}

\theoremstyle{plain}

\theoremstyle{definition}

\theoremstyle{remark}

\usepackage[textsize=tiny]{todonotes}

\icmltitlerunning{Offline Preference-Based Apprenticeship Learning}

\begin{document}

\twocolumn[
\icmltitle{Offline Preference-Based Apprenticeship Learning}



\icmlsetsymbol{equal}{*}

\begin{icmlauthorlist}
\icmlauthor{Daniel Shin}{yyy}
\icmlauthor{Daniel S. Brown}{yyy}
\icmlauthor{Anca D. Dragan}{yyy}
\end{icmlauthorlist}

\icmlaffiliation{yyy}{EECS Department, University of California, Berkeley}

\icmlcorrespondingauthor{Daniel Shin}{danielshin@berkeley.edu}
\icmlcorrespondingauthor{Daniel S. Brown}{dsbrown@berkeley.edu}

\icmlkeywords{Machine Learning, ICML}

\vskip 0.3in
]



\printAffiliationsAndNotice{}  

\begin{abstract}
Learning a reward function from human preferences is challenging as it typically requires having a high-fidelity simulator or using expensive and potentially unsafe actual physical rollouts in the environment. However, in many tasks the agent might have access to offline data from related tasks in the same target environment. While offline data is increasingly being used to aid policy optimization via offline RL, our observation is that it can be a surprisingly rich source of information for preference learning as well. We propose an approach that uses an offline dataset to craft preference queries via pool-based active learning, learns a distribution over reward functions, and optimizes a corresponding policy via offline RL. Crucially, our proposed approach does not require actual physical rollouts or an accurate simulator for either the reward learning or policy optimization steps. To test our approach, we identify a subset of existing offline RL benchmarks that are well suited for offline reward learning and also propose new offline apprenticeship learning benchmarks which allow for more open-ended behaviors. Our empirical results suggest that combining offline RL with learned human preferences can enable an agent to learn to perform novel tasks that were not explicitly shown in the offline data. 
\end{abstract}

\section{Introduction}
\label{introduction}

\begin{figure}
    \centering
    \includegraphics[width=\columnwidth]{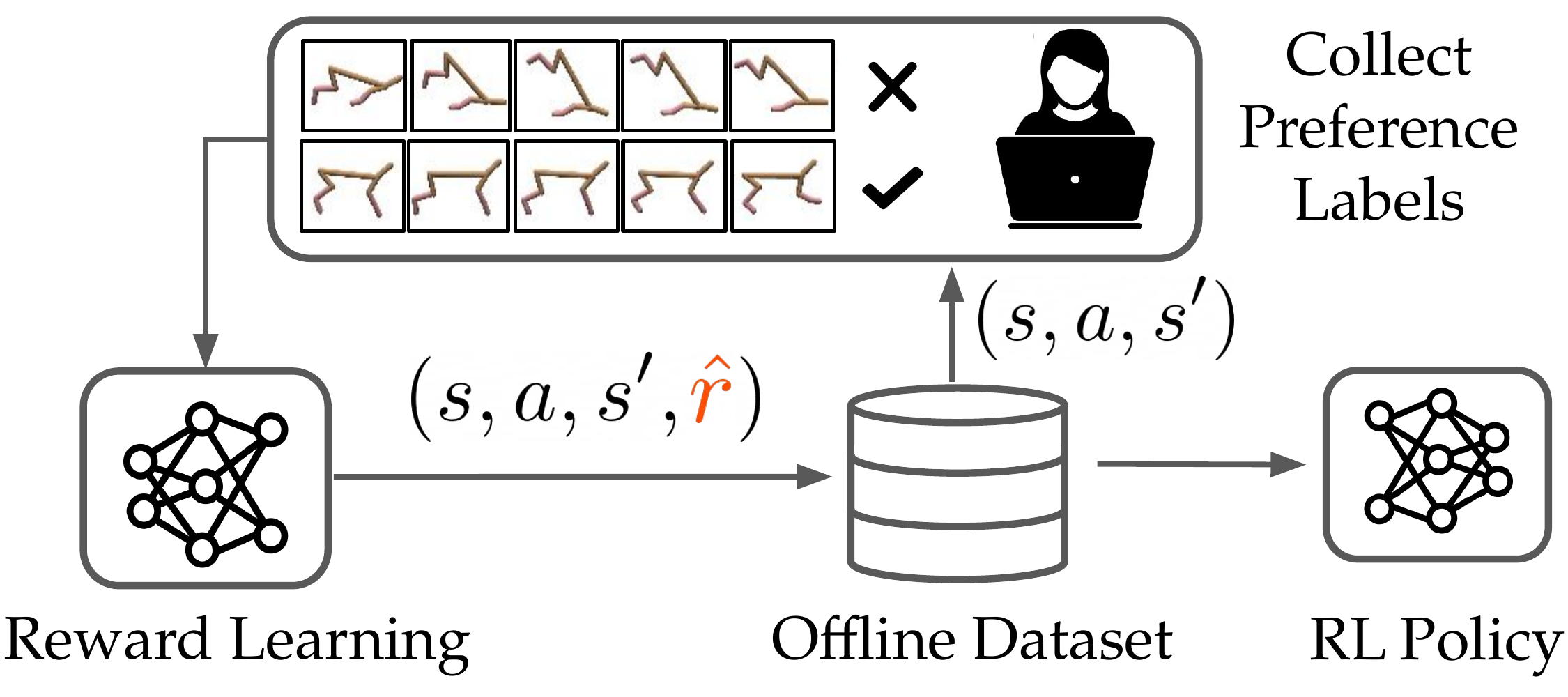}
    \caption{Offline Preference-Based Apprenticeship Learning (OPAL) enables safe and efficient preference-based policy customization without any environmental interactions. Given an offline database consisting of trajectories, OPAL queries an expert for preference labels over trajectory segments from the database, learns a reward function from preferences, and then performs offline RL using rewards provided by the learned reward function. 
    }
    \label{fig:opal_pipeline}
\end{figure}

\begin{figure*}[!th]
    \centering
    \includegraphics[width=\linewidth]{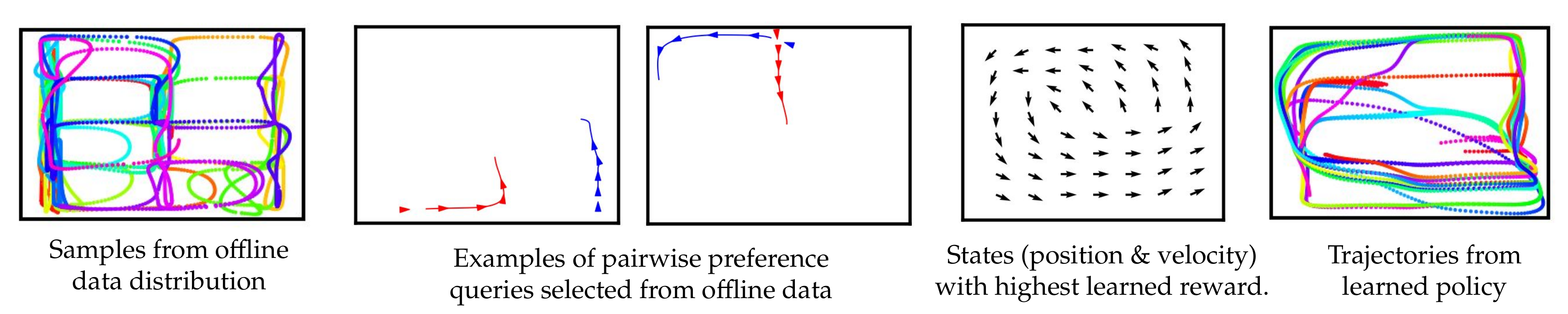}
    \caption{OPAL selects active queries from an offline dataset consisting of random point to point navigation. These active queries elicit preferences about a human's preferences (the human wants the robot to perform counter clockwise orbits and prefers blue over red trajectories). The learned reward function matches the human's preferences and, when combined with offline RL, leads to an appropriate policy that looks significantly different from the original offline data distribution. Offline data is often not collected for the specific task you want, but for other tasks. Thus, being able to repurpose data from a variety of sources is important for generalizing to different user preferences in offline settings where we can’t easily just gather lots of new online data.}
    \label{fig:active_qualitative}
\end{figure*}

For automated sequential decision making systems to effectively interact with humans in the real world, these systems need to be able to safely and efficiently adapt to and learn from different users. Apprenticeship learning~\cite{abbeel2004apprenticeship}---also called learning from demonstrations~\cite{argall2009survey} or imitation learning~\cite{osa2018algorithmic}---seeks to allow robots and other autonomous systems to learn how to perform a task through human feedback. 
Even though traditional apprenticeship learning uses expert demonstrations \citep{argall2009survey,osa2018algorithmic,arora2021survey}, preference queries---where a user is asked to compare two trajectory segments---have been shown to more accurately identify the reward \citep{jeon2020reward} while reducing the burden on the user to generate near-optimal actions~\cite{wirth2017survey,christiano2017deep}. However, one challenge is that asking these queries requires either executing them in the physical world (which can be unsafe), or executing them in a simulator and showing them to the human (which requires simulating the world with high enough fidelity). 

Learning a policy via RL has the same challenges of needing a simulator or physical rollouts~\cite{sutton2018reinforcement}. To deal with these problems, offline RL has emerged as a promising way to do policy optimization with access to \emph{offline} data only~\cite{levine2020offline}. Our key observation in this paper is that an analogous approach can be applied to reward learning: rather than synthesizing and executing preference queries in a simulator or in the physical world, we draw on an offline dataset to extract segments that would be informative for the user to compare. Since these segments have already been executed and recorded by the agent, all we need to do is replay them to the human, and ask for a comparison.



We call this approach Offline Preference-based Apprenticeship Learning (OPAL), a novel approach that consists of offline preference-based reward learning combined with offline RL. Our approach is summarized in Figure~\ref{fig:opal_pipeline}. OPAL has two appealing and desirable properties: safety and efficiency. No interactions with the environment are needed for either reward learning or policy learning, removing the sample complexity and safety concerns that come from trial and error learning in the environment. 

Having access to an offline dataset enables robots and other autonomous agents to reuse prior data that may come from a variety of sources, but may not be customized to a particular user's intent or preferences. Being able to repurpose data from a variety of sources is important for generalizing to different user preferences in offline settings where gathering new online data is difficult. Being able to use suboptimal offline data for the intended task is also a important in offline settings where optimal data may not be available.
Our results suggest that having a diverse offline dataset provides a surprising amount of flexibility.
By actively querying for comparisons between \emph{segments} (or snippets) from a variety of different rollouts, the agent can gain valuable information about how to customize it's behavior based on a user's preferences. This is true even when none of the data in the offline dataset explicitly demonstrates the desired behavior.

To effectively study offline reward learning, we require a set of benchmark domains to evaluate different approaches. 
However, while there are many standard RL benchmarks, there are many fewer benchmarks for reward learning or even imitation learning more broadly. Recent work has shown that simply using standard RL benchmarks and masking the rewards is not sufficiently challenging since often learning a +1 or -1 reward everywhere is sufficient for imitating RL policies~\cite{freire2020derail}. While there has been progress in developing imitation learning benchmarks~\cite{toyer2020magical,freire2020derail}, existing domains assume an online policy optimization setting, with full access to the environment which is only realistic in simulation due to efficiency and safety concerns. 

One of our contributions is to evaluate a variety of existing offline RL benchmarks~\cite{fu2020d4rl} in the offline apprenticeship learning setting, where we remove access to the true reward function. Surprisingly, we find that many offline RL benchmarks are ill-suited for reward learning---simply replacing all actual rewards in the offline dataset with zeros, or a constant, results in performance similar to, or better than, the performance obtained using the true rewards!  
Importantly, we identify a subset environments where simply using a trivial reward function guess results in significant degradation in policy performance, motivating the use of these environments when evaluating reward learning methods. We identify high offline data variability and multi-task data as two settings where reward learning is necessary for good policy learning. 

Because most existing offline RL domains are not suitable for studying the effects of reward learning, we also propose several new tasks designed for open-ended reward learning in offline settings.
Figure~\ref{fig:active_qualitative} shows an example from our experiments where the learning agent has access to a dataset composed of trajectories generated offline by a force-controlled pointmass robot navigating between random start and goal locations chosen along a 5x3 discrete grid. OPAL uses this dataset to learn to perform counter-clockwise orbits, a behavior never explicitly seen in the offline data. Figure~\ref{fig:active_qualitative} shows a sequence of active queries over trajectory snippets taken from the offline dataset. The blue trajectories were labeled by the human as preferable to the red trajectories. We also visualize the learned reward function. Using a small number of queries and a dataset generated by a very different process than the desired task, we are able learn the novel orbiting behavior in a completely offline manner.

We summarize the contributions of this paper as follows:
\begin{enumerate}
    \item We propose and formalize the novel problem of offline preference-based apprenticeship learning.
    \item We evaluate a large set of existing offline RL benchmarks, identify a subset that is well-suited for evaluating offline apprenticeship learning, and propose new benchmarks that allow for more open-ended behavior.
    \item  We perform an empirical comparison of different methods for maintaining uncertainty and generating active queries and provide evidence that ensemble-based disagreement queries outperform other approaches.
    \item Our results suggest that there is surprising value in offline datasets. The combination of offline reward learning and offline RL leads to highly efficient reward inference and enables agents to learn to perform tasks not explicitly demonstrated in the offline dataset.
\end{enumerate}

\section{Related Work}
Prior work on preference-based apprenticeship learning typically focuses on online methods that require actual rollouts in the environment or access to an accurate simulator or model~\cite{wirth2017survey,christiano2017deep,sadigh2017active,biyik2018batch,lee2021pebble}. However, accurate simulations and model-based planning are often not possible. Furthermore, even when an accurate simulator or model is available, there is the added burden of solving a difficult sim-to-real transfer problem~\cite{tobin2017domain,peng2018sim,chebotar2019closing}. In most real-world scenarios, the standard preference learning approach involving many episodes of trial and error in the environment is likely to be unacceptable due to safety and efficiency concerns.

Prior work on safe apprenticeship learning enables learners to estimate risky actions~\cite{zhang2016query,hoque2021thriftydagger} and request human assistance, optimizes policies for tail risk rather than expected return~\cite{lacotte2019risk,brown2020bayesian}, and provides high-confidence bounds on the performance of the imitation learner's policy~\cite{brown2020safe}; however, these methods all rely on either an accurate dynamics model or direct interactions with the environment. By contrast, our approach towards safety is to develop a fully offline apprenticeship learning algorithm to avoid costly and potentially unsafe physical data collection during reward and policy learning.

While there has been some work on offline apprenticeship learning, prior work has focused on simple environments with discrete actions and hand-crafted reward features~\cite{klein2011batch,bica2021learning} and requires datasets that consist of expert demonstrations~\cite{lee2019truly}. Other work has considered higher-dimensional continuous tasks, but assumes access to expert demonstrations or requires experts to label trajectories with explicit reward values~\cite{cabi2019scaling,zolna2020offline}. By contrast, we focus on fully offline apprenticeship learning via small numbers of preference queries which are typically much easier to provide than fine-grained reward labels or near-optimal demonstrations~\cite{saaty2008relative,wirth2017survey}.

There has been previous work that focused on apprenticeship learning from heterogeneous human demonstrations for resource scheduling and coordinating~\cite{paleja2019interpretable}. Our approach also works well with heterogeneous data but uses preferences to learn a reward function which is then used for offline RL algorithms instead of learning from demonstrations directly. Additionally, our approach is not limited to resource scheduling and is flexible in terms of domain applications.

\section{Problem Definition}
We model our problem as a Markov decision process (MDP), defined by the tuple $(S, A, r, P, \rho_0, \gamma)$, where $S$ denotes the state space, $A$ denotes the action space, $r: S\times A \rightarrow \mathbb{R}$ denotes the reward, $P(s' | s, a)$ denotes the transition dynamics, $\rho_0(s)$ denotes the initial state distribution, and $\gamma \in (0, 1)$ denotes the discount factor. 




In contrast to standard RL, we do not assume access to the reward function $r$. Furthermore, we do not assume access to the MDP during training. Instead, we are provided a static dataset, $\mathcal{D}$, consisting of trajectories, $\mathcal{D} = \{\tau_0, ..., \tau_N\}$, where each trajectory $\tau_i$ consists of a contiguous sequence of state, action, next-state transitions tuples $\tau_i = \{(s_{i,0}, a_{i,0}, s'_{i,0}), ..., (s_{i,M}, a_{i,M}, s'_{i,M})\}$. Unlike imitation learning, we do not assume that this dataset comes from a single expert attempting to optimize a specific reward function $r(s, a)$. Instead, the dataset $\mathcal{D}$ may contain data collected randomly, data collected from a variety of policies, or even from a variety of demonstrators.

Instead of having access to the reward function, $r$, we assume access to an expert that can provide a small number of pairwise preferences over trajectory snippets from an offline dataset $\mathcal{D}$. Given these preferences, the goal is to find a policy $\pi(a | s)$ that maximizes the expected cumulative discounted rewards (also known as the discounted returns), $J(\pi) = \mathbb{E}_{\pi, P, \rho_0} \left[ \sum_{t=0}^{\infty} \gamma^t r(s_t, a_t)\right]$, under the unknown true reward function $r$ in a fully offline fashion---we assume no access to the MDP other than the offline trajectories contained in $\mathcal{D}$.

\section{Offline Preference-Based Apprenticeship Learning}
We now discuss our proposed algorithm, Offline Preference-based Appenticeship Learning (OPAL). OPAL is an active preference-based learning approach which first sequentially selects new queries to be labeled by the human expert in order to learn a reward function that models the user's preferences and can be used to generate a customized policy based on a user's preferences via offline RL.


As established by a large body of prior work~\cite{christiano2017deep,ibarz2018reward,brown2019extrapolating,palan2019learning}, the Bradley-Terry pairwise preference model~\cite{bradley1952rank} is an appropriate model for learning reward functions from user preferences over trajectories. Given a preference over trajectories, $\tau_i \prec \tau_j$, we seek to maximize the probability of the preference label:
\begin{equation}
P \big(\tau_i \prec \tau_j \mid \theta) =  \frac{\exp \displaystyle\sum_{s \in \tau_j} \hat{r}_\theta(s)}{\exp \displaystyle\sum_{s \in \tau_i} \hat{r}_\theta(s) + \exp \displaystyle\sum_{s \in \tau_j} \hat{r}_\theta(s)},
\end{equation}
by approximating the reward at state $s$ using a neural network, $\hat{r}_{\theta}(s)$, such that $\sum_{s \in \tau_i} \hat{r}_{\theta}(s) < \sum_{s \in \tau_j} \hat{r}_{\theta}(s)$ when $\tau_i \prec \tau_j$.

OPAL actively queries to obtain informative preference labels over trajectory snippets sampled from the offline dataset. In this paper, we investigate two different methods to represent uncertainty (ensembles and Bayesian dropout) along with two different acquisition functions (disagreement and information gain). We describe these approaches below.

\subsection{Representing Reward Uncertainty}
We compare two of the most popular methods for obtaining uncertainty estimates when using deep learning: ensembles~\cite{lakshminarayanan2016simple} and Bayesian dropout~\cite{gal2016dropout}. On line 4 of Algorithm~\ref{alg:opal}, we initialize an ensemble of reward models for ensemble queries or a single reward model for Bayesian dropout. 

\paragraph{Ensemble Queries}
Following work on online active preference learning work by \citet{christiano2017deep}, we test the effectiveness of training an ensemble of reward models to approximate the reward function posterior using the Bradley-Terry preference model. Similar to prior work~\cite{christiano2017deep,reddy2020learning}, we found that initializing the ensemble networks with different seeds was sufficient to produce a diverse set of reward functions. 

\paragraph{Bayesian Dropout}
We also test the effectiveness of using dropout to approximate the reward function posterior. As proposed by \citet{gal2016dropout}, we train a reward network using pairwise preferences and apply dropout to the last layer of the network during training. To predict a return distribution over candidate pairs of trajectories, we still apply dropout and pass each trajectory through the network multiple times to obtain a distribution over the returns.

\subsection{Active Learning Query Selection}
In contrast to prior work on active preference learning, we do not require on-policy rollouts in the environment~\cite{christiano2017deep,lee2021pebble} nor require synthesizing queries using a model~\cite{sadigh2017active}. Instead, we generate candidate queries by randomly choosing pairs of sub-trajectories obtained from the offline dataset. 
Given a distribution over likely returns for each trajectory snippet in a candidate preference query, we estimate the value of obtaining a label for this candidate query.
We consider two methods for computing the value of a query: disagreement and information gain. We then take the trajectory pair with the highest predicted value and ask for a pairwise preference label.
On line 6 of Algorithm~\ref{alg:opal}, we can either use disagreement or information gain as the estimated value of information (VOI).

\paragraph{Disagreement}
When using disagreement to select active queries, we select pairs with the highest ensemble disagreement among the different return estimates obtained from either ensembling or Bayesian dropout. Following \citet{christiano2017deep}, we calculate disagreement as the variance in the binary comparison predictions: if fraction $p$ of the posterior samples predict $\tau_i 	\succ \tau_j$ while the other $1-p$ ensemble models predict $\tau_i \preceq \tau_j$, then the variance of the query pair $(\tau_i, \tau_j)$ is $p(1-p)$.

\paragraph{Information Gain Queries}
As an alternative to disagreement, we also consider the expected information gain~\cite{cover1999elements} between the reward function parameters $\theta$ and the outcome $Y$ of querying the human for a preference. We model our uncertainty using an approximate posterior $p(\theta \mid \mathcal{D})$, given by training an ensemble or dropout network on our offline dataset $\mathcal{D}$. 
\citet{houlsby2011bayesian} show that the information gain of a potential query can be formulated as:
\begin{equation}
I(\theta; Y \mid \mathcal{D}) = H(Y \mid \mathcal{D}) - \mathbb{E}_{\theta \sim p(\theta \mid \mathcal{D})}[H(Y \mid \theta, \mathcal{D})].
\end{equation}
Intuitively, the information gain will be maximized when the first term is high, meaning that the overall model has high entropy, but the second term is low, meaning that each individual hypothesis $\theta$ from the posterior assigns low entropy to the outcome $Y$. This will happen when the individual hypotheses strongly disagree with each other and there is no clear majority.
We approximate both terms in the information gain equation with samples obtained via ensemble or dropout. See Appendix~\ref{app:info_gain} for further details.

\subsection{Policy Optimization}
Given a learned reward function obtained via preference-learning, we can then use any existing offline or batch RL algorithm~\cite{levine2020offline} to learn a policy without requiring knowledge of the transition dynamics or rollouts in the actual environment. 

\section{Experiments and Results}
We first evaluate a variety of popular offline RL benchmarks from D4RL~\cite{fu2020d4rl} to determine which domains are most suited for evaluating offline apprenticeship learning. Prior work on reward learning has shown that simply showing good imitation learning performance on an RL benchmark is not sufficient to demonstrate good reward learning~\cite{freire2020derail}. In particular, we seek domains where simply using an all zero or constant reward does not result in good performance. After isolating several domains where learning a shaped reward actually matters (Section~\ref{sec:eval_benchmarks}), we evaluate OPAL on these selected benchmark domains and investigate the performance of different active query strategies (Section~\ref{sec:subset_eval}). We finally, propose and evaluate several tasks specifically designed for offline apprenticeship learning (Section~\ref{sec:new_domains}).

\begin{figure*}
     \centering
     \begin{subfigure}[b]{0.18\textwidth}
         \centering
         \includegraphics[width=\textwidth]{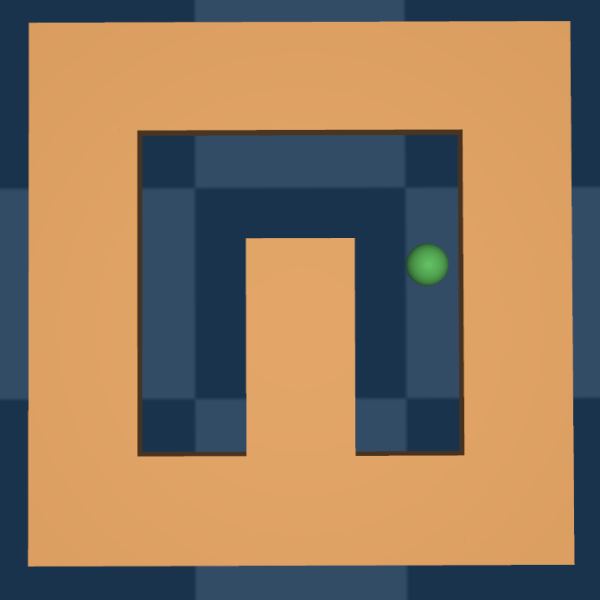}
         \caption{U-Maze}
         \label{fig:umaze}
     \end{subfigure}
     \hfill
     \begin{subfigure}[b]{0.18\textwidth}
         \centering
         \includegraphics[width=\textwidth]{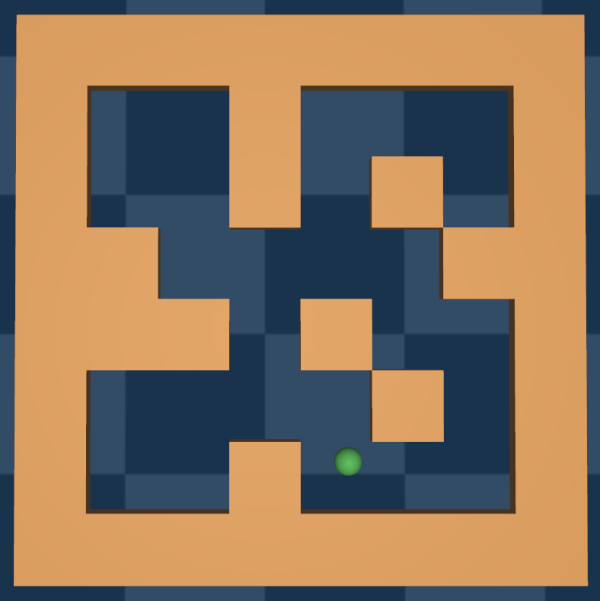}
         \caption{Medium Maze}
         \label{fig:medium_maze}
     \end{subfigure}
     \hfill
          \begin{subfigure}[b]{0.18\textwidth}
         \centering
         \includegraphics[width=\textwidth]{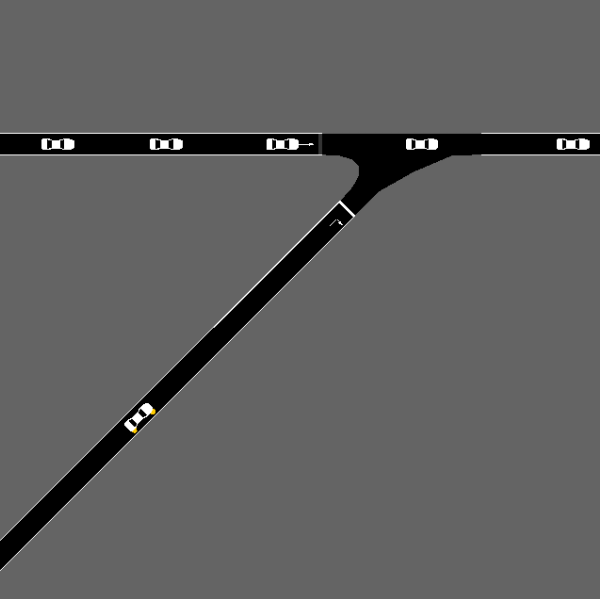}
         \caption{Flow Merge}
         \label{fig:flow_merge}
     \end{subfigure}
     \hfill
     \begin{subfigure}[b]{0.18\textwidth}
         \centering
         \includegraphics[width=\textwidth]{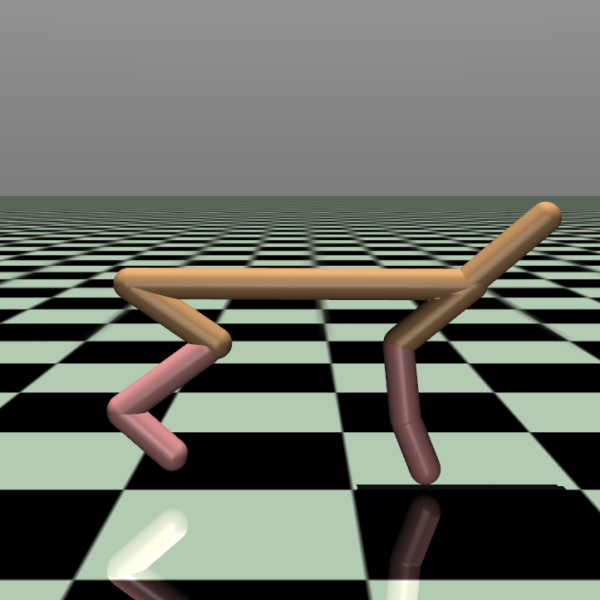}
         \caption{Halfcheetah}
         \label{fig:halfcheetah}
     \end{subfigure}
     \hfill
\begin{subfigure}[b]{0.18\textwidth}
         \centering
         \includegraphics[width=\textwidth]{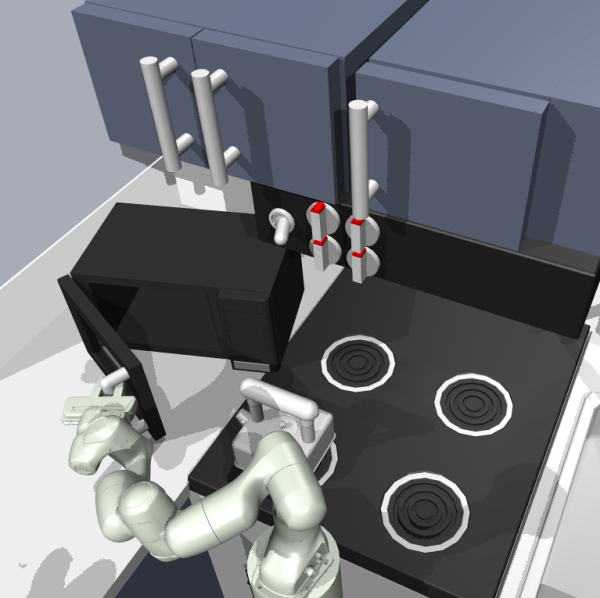}
         \caption{Franka Kitchen}
         \label{fig:kitchen-complete}
     \end{subfigure}
        \caption{Experimental domains chosen from D4RL~\cite{fu2020d4rl} for use in offline preference-based apprenticeship learning.}
        \label{fig:d4rl_domains}
\end{figure*}

\subsection{Evaluating Offline RL Benchmarks}\label{sec:eval_benchmarks}


An important assumption made in our problem statement is that we do not have access to the reward function. Before describing our approach for reward learning, we first investigate in what cases we actually need to learn a reward function. We study a set of popular benchmarks used for offline RL~\cite{fu2020d4rl}. To test the sensitivity of these methods to the reward function, we evaluate the performance of offline RL algorithms on these benchmarks when we set the reward equal to zero for each transition in the offline dataset and when we set all the rewards equal to a constant (the average reward over the entire dataset). 

We evaluated four of the most commonly used offline RL algorithms: Advantage Weighted Regression (AWR)~\cite{peng2019awr}, Batch-Constrained deep Q-learning (BCQ)~\cite{fujimoto2019bcq}, Bootstrapping Error Accumulation Reduction (BEAR)~\cite{kumar2019stabilizing}, and Conservative Q-Learning (CQL)~\cite{kumar2020cql}.
Table~\ref{tab:finalreturn_ablation} shows the resulting performance across a large number of tasks from the D4RL benchmarks~\cite{fu2020d4rl}. 
We find that for tasks with only expert data, there is no need to learn a reward function since using a trivial reward function often leads to expert performance. We attribute this to the fact that many offline RL algorithms are similar to behavioral cloning~\cite{levine2020offline}. Thus, when an offline RL benchmarks consists of only expert data, offline RL algorithms do not need reward information to perform well. In Appendix~\ref{app:no_reward} we show that AWR reduces to behavioral cloning in the case of an all zero reward function. 

Conversely, when the offline dataset contains a wide variety of data that is either random or multi-task (e.g. the Maze2D environment consists of an agent traveling to randomly generated goal locations), we find that running offline RL algorithms on the dataset with an all zero or constant reward function significantly degrades performance.
\begin{table*}[t]
\caption{Offline RL performance on D4RL benchmark tasks comparing the performance with the true reward to performance using a constant reward equal to the average reward over the dataset (Avg) and zero rewards everywhere (Zero). Even when all the rewards are set to the average or to zero, many offline RL algorithms still perform surprisingly well. In this table, we present the experiments ran with AWR. Results are averages over three random seeds. Bolded environments are ones where we find the degradation percentage to be over a chosen threshold and are used for later experiments with OPAL. The degradation percentage is calculated as ${\max(\text{GT} - \max(\text{AVG}, \text{ZERO}, \text{RANDOM}), 0)\over |\text{GT}| }\times 100\%$
and the threshold is set to 25\%. The degradation percentage is used to determine domains where trivial reward functions do not lead to good performance to isolate the effect of reward learning in later experiments.}
\label{tab:finalreturn_ablation}
\vskip 0.15in
\begin{center}
\begin{small}
\begin{sc}
\begin{tabular}{lrrrrrr}
\toprule
Task &  
GT &  Avg & Zero & Random & Degradation \% \\
\midrule
    flow-ring-random-v1 & -42.5 & -42.9 & -44.3 & -166.2 & 0.94\\
    \textbf{flow-merge-random-v1} & 160.3 & 85.2 & 85.6 & 117.0	& 27.0 \\
    \midrule
    \textbf{maze2d-umaze} & 104.4 & 56.5 & 55.0 & 49.5 & 45.9 \\
    \textbf{maze2d-medium} & 134.6 & 30.5 & 34.5 & 44.8 & 66.7 \\
    \midrule
  \textbf{halfcheetah-random} & 11.0 & -49.4 & -48.3 & -285.8 & 539.1 \\
    halfcheetah-medium-replay & 4138.2 & 3934.7 & 3830.2 & -285.8 & 4.9\\
    halfcheetah-medium & 4096.0 & 3984.4 & 3898.3 & -285.8 & 2.7 \\
    halfcheetah-medium-expert & 669.9 & 401.7 & 560.5 & -285.8 &  16.3\\
    halfcheetah-expert & 467.8 & 511.1 & 493.1 & -285.8 & 0.0\\
    \midrule
    hopper-random & 123.3 & 129.8 & 29.1 & 18.3 & 0.0 \\
    hopper-medium-replay & 1045.6 & 1127.4 & 954.6 & 18.3 & 0.0 \\
    hopper-medium & 1152.1 & 1192.8	& 963.6 & 18.3 & 0.0 \\
    hopper-medium-expert & 623 & 615.8 & 587.3 & 18.3 & 1.2 \\
    hopper-expert & 571.7 & 617.2 & 427.6 & 18.3 & 0.0 \\
    \midrule
    \textbf{kitchen-complete} & 0.3	& 0.2 & 0.3 & 0.0 & 25.0 \\
    kitchen-mixed & 0.3 & 0.5 & 0.3 & 0.0 & 16.7 \\
    kitchen-partial & 0.3 & 0.2 & 0.3 & 0.0 & 0.0 \\
\bottomrule
\end{tabular}
\end{sc}
\end{small}
\end{center}
\vskip -0.1in
\end{table*}

\subsection{Apprenticeship Learning on a Subset of D4RL}\label{sec:subset_eval}
To evaluate the benefit of offline apprenticeship learning, we focus our attention on the D4RL benchmark domains shown in Figure~\ref{fig:d4rl_domains} which showed significant degradation in Table~\ref{tab:finalreturn_ablation} since these are domains where we can be confident that good performance is due to learning a good reward function. We define significant degradation to be degradation percentage over 25\%, which was chosen to give enough room for OPAL to improve on in comparison to uninformative rewards like avg or zero masking. We selected both of the Maze environments, the Flow Merge traffic simulation, the  Halfcheetah random environment, and the Franka Kitchen-Complete environment. 
To facilitate a better comparison across different methods for active learning, we use oracle preference labels, where one trajectory sequence is preferred over another if it has higher ground-truth return. 
We report the performance of OPAL using AWR~\cite{peng2019awr}, which we empirically found to work for policy optimization across the different tasks. In the appendix we also report results when using CQL~\cite{kumar2020cql} for policy optimization.

\subsubsection{Maze Navigation}
We first consider the Maze2d domain \cite{fu2020d4rl}, which involves moving a force-actuated ball (along the X and Y axis) to a fixed target location. The observation is 4 dimensional, which consists of the $(x, y)$ location and velocities. The offline dataset consists of one continuous trajectory of the agent navigation to random intermediate goal locations. The true reward, which we only use for providing synthetic preference labels, is the negative exponentiated distance to a held-out goal location. 

For our experimental setup, we first randomly select 5 pairs of trajectory snippets and train 5 epochs with our models. After this initial training process, for each round, one additional pair of trajectories is queried to be added to the training set and we train one more epoch on this augmented dataset. 
The learned reward model is then used to predict the reward labels for all the state transitions in the offline dataset, which is then used to train a policy via offline RL (e.g. AWR \cite{peng2019awr}). As a baseline, we also compare against
a random, non-active query baseline.


\begin{figure*}[t]
     \centering
     \
     \begin{subfigure}[b]{0.3\textwidth}
         \centering
         \includegraphics[width=\textwidth]{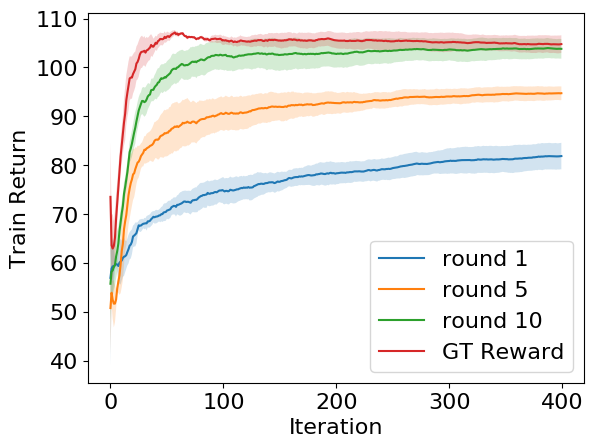}
         \caption{Maze2D-Umaze}
     \end{subfigure}
      \hfill
     \begin{subfigure}[b]{0.3\textwidth}
         \centering
         \includegraphics[width = \textwidth]{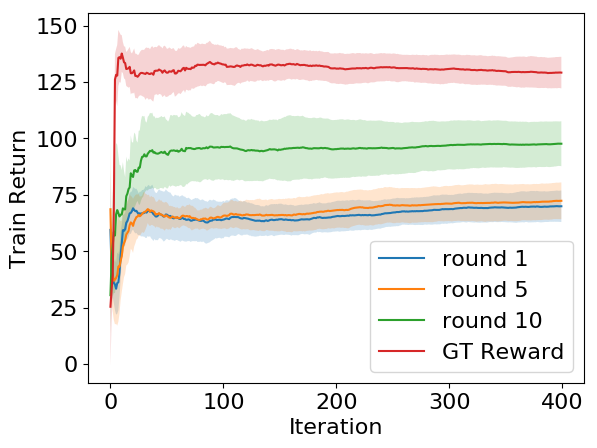}
        \caption{Maze2D-Medium}
     \end{subfigure}
     \hfill
     \begin{subfigure}[b]{0.3\textwidth}
         \centering
         \includegraphics[width = \textwidth]{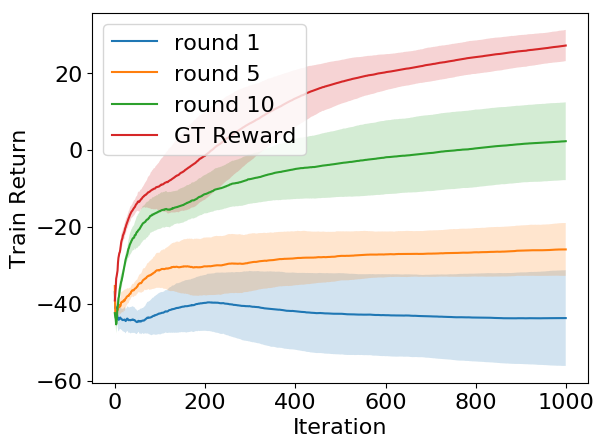}
         \caption{Half-Cheetah-v2}
     \end{subfigure}
     \caption{(a) Ensemble disagreement after 10 rounds has similar performance to ground truth reward in Maze2d-Umaze. (b) Ensemble disagreement after 10 rounds has slightly worse performance compared to ground truth reward in Maze2d-Medium. (c) Ensemble disagreement after 10 rounds has similar performance compared to ground truth reward in Halfcheetah.}
     \label{fig:maze2d_mujoco}
\end{figure*}

The results are provided in Table~\ref{tab:finalreturn}.
We found that ensemble disagreement is the best performing approach for both Maze2D-Umaze and Maze2D-Medium and outperforms T-REX. 
The offline RL performance for ensemble disagreement are shown in Figure~\ref{fig:maze2d_mujoco}. We find that ensemble disagreement on Maze2D-Umaze is able to quickly approach the performance when using the ground-truth reward function on the full dataset of 1 million state transitions after only 15 preference queries (5 initial + 10 active). Maze2D-Medium is more difficult and we find that 15 preference queries is unable to recover a policy on par with the policy trained with ground truth rewards. 
We hypothesize that this is due to the fact that Maze2D is goal oriented and learning a reward with trajectory comparisons might be difficult since the destination matters more than the path. 

\begin{table*}[t]
\caption{\textbf{OPAL Performance Using Reward Predicted After N Rounds of Querying}: Offline RL performance using rewards predicted either using a static number of randomly selected pairwise preferences (T-REX)~\cite{brown2019extrapolating}, or active queries using Ensemble Disagreement(EnsemDis), Ensemble Information Gain (EnsemInfo), Dropout Disagreement (DropDis), and Dropout Information Gain (DropInfo). N is set to 5 for Maze2D-Umaze, 10 for Maze2d-Medium and flow-merge-random, 15 for Halfcheetah. N is selected based on the size of the dimension of observation space and the complexity of the environment. Results are averages over three random seeds and are normalized so the performance of offline RL using the ground-truth rewards is 100.0 and performance of a random policy is 0.0. 
} 
\label{tab:finalreturn}
\vskip 0.15in
\begin{center}
\begin{small}
\begin{sc}
\begin{tabular}{l|rrrrr}
\toprule
& & \multicolumn{4}{c}{Query Aquisition Method} \\
& Random &  EnsemDis & EnsemInfo & DropDis & DropInfo\\
\midrule
maze2d-umaze & 78.2 & \textbf{93.5} & 89.2 & 88.0 & 61.3 \\
maze2d-medium & 71.6 & \textbf{86.4} & 71.0 & 52.6 & 44.4 \\
halfcheetah & 96.1 & \textbf{113.7} & 100.6 & 84.4 & 91.7 \\
flow-merge-random & \textbf{110.1} & 89.0 & 92.1 & 86.2 & 84.2 \\
kitchen-complete & 79.6 & 105.0 & \textbf{158.8} & 48.5 & 65.4  \\
\bottomrule
\end{tabular}
\end{sc}
\end{small}
\end{center}
\vskip -0.1in
\end{table*}

\subsubsection{MuJoCo Tasks}
We next tested OPAL on the MuJoCo environment Halfcheetah-v2. 
For the tasks, the agent is rewarded for making the Halfcheetah move forward as fast as possible. The MuJoCo tasks presents an additional challenge in comparison to the Maze2d domain since they have higher dimensional observations spaces, 17 for Halfcheetah-v2. The dataset contains 1 million transitions obtained by rolling out a random policy. Our experimental setup is similar to Maze2D, except we start with 50 pairs of trajectories instead of 5 and we add 10 trajectories per active query instead of one. We found that increasing the number of queries was helpful due to the fact that MuJoCo tasks have higher dimensional observational spaces, which requires more data points to learn an accurate reward function.
The learning curve is shown in Figure~\ref{fig:maze2d_mujoco}~(c).
We found ensemble disagreement to be the best performing approach for the Halfcheetah task. Notably, the policy learned after 10 rounds of querying is on par with the policy learned with the the ground truth reward. 


\subsubsection{Flow Merge}
The Flow Merge environment involves directing traffic such that the flow of traffic is maximized in a highway merging scenario. Results are shown in Table~\ref{tab:finalreturn}. Random queries achieved the best performance, but all query methods were able to recover near ground truth performance. 

\subsubsection{Robotic Manipulation}

The Franka Kitchen domain involves interacting with various objects in the kitchen to reach a certain state configuration. The objects you can interact with includes a water kettle, a light switch, a microwave, and cabinets. Our results in Table~\ref{tab:finalreturn_ablation} show that setting the reward to all zero or a constant reward causes the performance to degrade significantly.
Interestingly, the results in Table~\ref{tab:finalreturn} show that OPAL using Ensemble Disagreement and Ensemble InfoGain are able to shape the reward in a way that actually leads to better performance than using the ground-truth reward for the task. In the appendix, we plot the learning curves and find that OPAL also converges to a better performance faster. 

\subsection{New Offline Preference-Based Apprenticeship Learning Tasks}\label{sec:new_domains}
To explore the full benefits of learning a reward function from preferences, we propose and study several new tasks to highlight the need for a shaped reward function, rather than simply a goal indicator. To create these environments we adapted common gym environments including several that are in the D4RL set of benchmarks.

\subsubsection{Maze Navigation with Constraint Region}
\begin{figure}[t!]
     \centering
         \includegraphics[width = 0.4\linewidth]{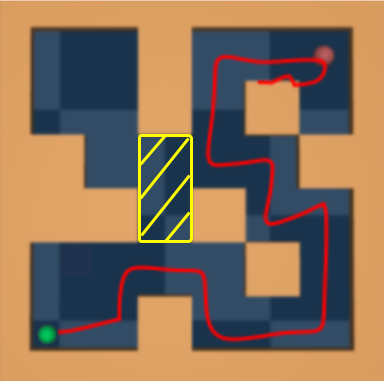}
    \caption{\textbf{Constrained Goal Navigation}. The highlighted yellow region represents a constraint region that the human prefers the agent to avoid while also traveling to the goal position shown in red. OPAL produces trajectories that match this preference by taking a more round-about, but more preferred, path to the goal.}
     \label{fig:maze_constraint}
     
\end{figure}

We created a new variant of the Maze2d-Medium task where there is a constraint region in the middle of the maze that the supervisor does not want the robot to enter. Figure~\ref{fig:maze_constraint} shows this scenario where the entering the highlighted yellow region is undesirable. After only 25 active queries using ensemble disagreement we obtain the behavior shown in Figure~\ref{fig:maze_constraint} where the offline RL policy has learned to reach the goal while avoiding the constraint region. 

\subsubsection{Open Maze Behaviors}
\begin{figure}[t]
     \centering
     \begin{subfigure}[b]{0.23\linewidth}
         \centering
         \includegraphics[width=\linewidth]{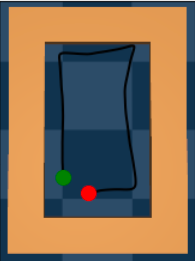}
     \end{subfigure}
     \hfill
     \begin{subfigure}[b]{0.23\linewidth}
         \centering
         \includegraphics[width = \linewidth]{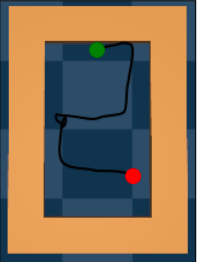}
     \end{subfigure}
     \hfill
     \begin{subfigure}[b]{0.23\linewidth}
         \centering
         \includegraphics[width=\linewidth]{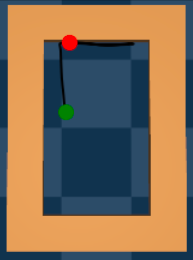}
     \end{subfigure}
     \hfill    
     \begin{subfigure}[b]{0.23\linewidth}
         \centering
         \includegraphics[width=\linewidth]{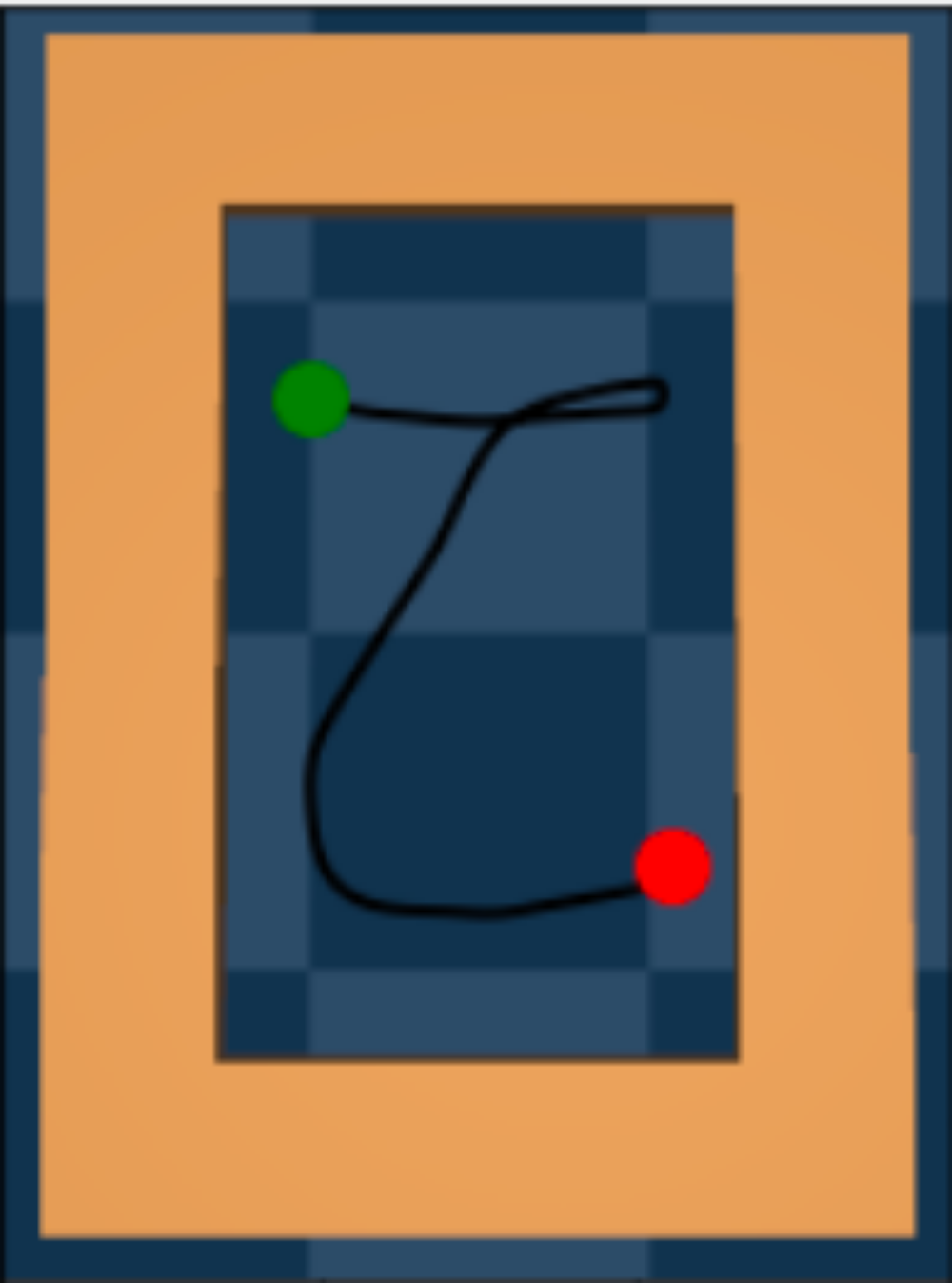}
     \end{subfigure}
    \caption{\textbf{Learned ``Shape" Policies} adapt to particular user's preferences to learn clockwise orbits, a digital 2 shape, preferring to stay at the top, and a ``Z" shape, respectively. Green circles represent random initial states and red circles represent the position of the agent at the end of a rollout.}
     \label{fig:open_maze_variety}
\end{figure}

Next, we took the D4RL Open Maze environment and dataset and used it to teach an agent to patrol the domain in counter clockwise orbits, clockwise orbits, approximating a digital 2 shape, preferring to stay at the top, and approximating a "Z" shape. The open maze counter close wise orbiting uses human preferences provided by one of the authors by showing the human user trajectory snippets visualized as in~\ref{fig:active_qualitative} and querying preferences. The dataset only contains the agent moving to randomly chosen goal locations, thus this domain highlights the benefits of stitching together data from an offline database as well as the benefits of learning a shaped reward function rather than simply using a goal classifier to use as the reward function which has been used in prior work on offline RL work~\cite{eysenbach2021replacing}. The original data distribution, samples of the preference queries, and the resulting learned policies are shown in figures~\ref{fig:active_qualitative} and
\ref{fig:open_maze_variety}. Our results involving the open maze behaviors in Figure
\ref{fig:open_maze_variety} and our work on constrained maze in Figure \ref{fig:maze_constraint} can be seen as a mixture of experts since the data consists of a mixture of demonstrations for a variety of starts and goals. OPAL can leverage diverse data to achieve different preferences (including ones not explicitly demonstrated), such as taking a longer rather than shorter path to a goal because of a specific user constraint preference in the medium maze, or various behaviors in the open maze.



\subsubsection{Open Ended CartPole Behaviors}
We created an offline dataset of 1,000 random trajectories of length 200 in a modified CartPole domain where the trajectories only terminate at the end of 200 timesteps---this allows the pole to swing below the track and also allows the cart to move off the visible track to the left or right. Given this dataset of behavior agnostic data, we wanted to see whether we could optimize different policies, corresponding to different human preferences. We can also see this data from a random controller as highly suboptimal data for the following tasks. We optimized the following policies: \textit{balance}, where the supervisor prefers the pole to be balanced upright and prefers the cart to stay in the middle of the track; \textit{clockwise windmill}, where the supervisor prefers the cart to stay in the middle of the track but prefers the pole to swing around as fast as possible in the clockwise direction; and \textit{counterclockwise windmill}, which is identical to clockwise windmill except the preference is for the pole to swing in the counterclockwise direction. OPAL is able to learn policies for all three behaviors. Videos of learned behaviors are available at \url{https://sites.google.com/view/offline-prefs}.

\section{Discussion and Future Work}
Our goal is to learn a policy from user preferences without requiring access to a model, simulator, or interactions with the environment. Toward this goal we propose OPAL: Offline Preference-based Apprenticeship Learning.  We evaluate several different offline RL benchmarks and find that those that contain mainly random data or multi-task data are best suited for reward inference. While most prior work on apprenticeship learning has focused on learning from expert demonstrations, preference-based reward learning enables us to learn a user's intent even from random datasets by selecting informative sub-sequences of transitions and requesting pairwise preferences over these sub-sequences. Our results suggest that using ensemble disagreement as an acquisition function is the best choice since it leads to the best performance for 3 out of the 5 tasks and outperforms the random baseline for all 5 tasks.  Additionally, using ensemble disagreement with only 10-15 preference queries we are able to learn policies that perform similar to policies trained with full access to tens of thousands of ground-truth reward samples. Finally, our results suggest there is surprising value in offline datasets even if the datasets do not contain the data for the right tasks or if the datasets contain suboptimal data.


We are excited about the potential of offline reward and policy learning and believe that our work provides a stepping stone towards safer and more efficient autonomous systems that can better learn from and interact with humans. 
Preference-learning enables users to teach a system without needing to be able to directly control the system or demonstrate optimal behavior. We believe this makes preferences ideally suited for many tasks in as healthcare, finance, or robotics. Future work includes developing more complex and realistic datasets and benchmarks and developing offline RL algorithms and active query strategies that are well-suited for reward inference and offline RL.

\bibliography{opal_refs}
\bibliographystyle{icml2022}

\newpage
\appendix
\onecolumn

\clearpage
\appendix

\section{Information Gain}\label{app:info_gain}
As an alternative to disagreement, we also consider the expected information gain~\cite{cover1999elements} between the reward function parameters $\theta$ and the outcome $Y$ of querying the human for a preference. We model our uncertainty using an approximate posterior $p(\theta \mid \mathcal{D})$, given by training an ensemble or dropout network on our offline dataset $\mathcal{D}$. 
\citet{houlsby2011bayesian} show that the information gain of a potential query can be formulated as:
\begin{equation}
I(\theta; Y \mid \mathcal{D}) = H(Y \mid \mathcal{D}) - \mathbb{E}_{\theta \sim p(\theta \mid \mathcal{D})}[H(Y \mid \theta, \mathcal{D})].
\end{equation}
Intuitively, the information gain will be maximized when the first term is high, meaning that the overall model has high entropy, but the second term is low, meaning that each individual sample from the posterior has low entropy. This will happen when the individual hypotheses strongly disagree with each other and there is no clear majority.
We approximate both terms in the information gain equation with samples obtained via ensemble or dropout.

Given a pair of trajectories $(\tau_A, \tau_B)$, let $Y=0$ denote that the human prefers trajectory $A$ and $Y=1$ denote that the human prefers trajectory $B$. Use the Bradley-Terry preference model~\cite{bradley1952rank} we have that 
\begin{align} 
    P(Y = 0 \mid \theta, \mathcal{D}) = \frac{\exp(\beta R(\xi_A))}{\exp(\beta R(\xi_A)) + \exp(\beta R(\xi_B))}
\end{align}
where $R(\xi) = \sum_{(s,a) \in \tau_i} \hat{r}_{\theta}(s)$ and $P(Y = 1 \mid \theta, \mathcal{D}) = 1- P(Y = 0 \mid \theta, \mathcal{D})$.

To perform queries using Information Gain, we evaluate a pool of potential trajectory pairs, evaluate the information gain for each pair and query the human for the pair that has the highest information gain. 

\begin{equation}
\mathbb{E}_{\theta \sim p(\theta \mid \mathcal{D})}[H(Y \mid \theta, \mathcal{D})] \approx \frac{1}{M} \sum_{i = 1}^M H(Y \mid \theta_i, \mathcal{D}), 
\end{equation}
where $\theta_i, i = 1, \ldots, M$ are $M$ approximate posterior samples from $p(\theta \mid \mathcal{D})$ obtained from each member of the ensemble or from Bayesian dropout sampling.

Because we only consider pairwise preference queries, computing the entropy corresponds to
\begin{equation}
H(Y|\theta_i, \mathcal{D}) = - \sum_{y=0}^1 P(Y=y \mid \theta_i, \mathcal{D}) \log P(Y=y \mid \theta_i, \mathcal{D})
\end{equation}

To compute the first entropy term, we simply take the entropy of the average over the posterior as:
\begin{equation}
H(Y|\mathcal{D}) = - \sum_{y=0}^1 \bar{p}_y \log \bar{p}_y
\end{equation}
where $ \bar{p}_y = \frac{1}{M} \sum_{i=1}^M P(Y=y \mid \theta_i, \mathcal{D})$.

\section{Evaluating Offline RL Benchmarks}\label{app:no_reward}

    


We find that using average and zero masking are very competitive with using the true rewards on many of the benchmarks for D4RL and that these baselines often perform significantly better than a purely random policy.

This is a byproduct of offline reinforcement learning algorithms needing to learn in the presence of out-of-distribution actions. There are broadly two classes of methods towards solving the out-of-distribution problem. Behavioral cloning based methods like Advantage Weighted Regression used in our experiments, train only on actions observed in the dataset, which avoid OOD actions completely. 
Dynamic programming (DP) methods, like BCQ, constrains the trained policy distribution to lie close to the behavior policy that generated the dataset. As a result of offline reinforcement learning algorithms constraining action to be similar to that of the static dataset, if the static dataset consists of exclusively expert actions, then the offline reinforcement learning algorithm will recover a policy that has expert like action regardless of the reward.

\subsection{Advantage-Weighted Regression}
Consider Advantage-Weighted Regression with a constant reward function $r(s,a) = c$. The algorithm is simple. We have a replay buffer from which we sample transitions and estimate the value function via regression 
\begin{equation}
    V* \gets \arg \min_V \mathbb{E}_{s,a\sim D} \left[ \| \mathcal{R}^\mathcal{D}_{s,a} - V(s) \| \right]
\end{equation}
then the policy is updated via supervised learning:
\begin{equation}
    \pi \gets \arg \max_\pi \mathbb{E}_{s,a\sim D} \left[ \log \pi(a|s) \exp \left(\frac{1}{\beta} \big( \mathcal{R}^\mathcal{D}_{s,a} - V(s) \big) \right) \right]
\end{equation}

If the rewards are all zero, then we have $V(s) = 0$, $\forall s \in \mathcal{S}$ and 
\begin{align}
    \pi &\gets \arg \max_\pi \mathbb{E}_{s,a\sim D} \left[ \log \pi(a|s) \exp \left(\frac{1}{\beta} 0 \right) \right] \\
    &=  \max_\pi \mathbb{E}_{s,a\sim D} \left[ \log \pi(a|s)  \right].
\end{align}
This is exactly the behavioral cloning loss which will learn to take the actions in the replay buffer. Thus, AWR is identical to BC when the rewards are all zero.

For non-zero, constant rewards, the advantage term will be zero (at least for deterministic tasks). If all trajectories from the state $s$ have the same length, then we will have zero advantage. However, if there are trajectories that terminate earlier than others, then AWR will put weights on these trajectories proportional to their length (for positive rewards $c>0$) and inversely proportional to their length (for negative rewards $c<0$). Thus, a terminal state will leak information and a good policy can be learned without an informative reward function.

\section{Comparison of AWR and CQL for OPAL}
In Table~\ref{tab:finalreturn_awr_cql} we show results for OPAL when using AWR~\cite{peng2019awr} for policy optimization and when using CQL~\cite{kumar2020cql} for policy optimization. Interestingly, our results suggest that when using CQL, using dropout to represent uncertainty performs better than using an ensemble. However, when using AWR for policy optimization, ensembles perform better. We hypothesize that different query mechanisms can have complex interactions with the actual policy learning algorithm. Better understanding how active queries and offline reward learning affects the performance of different offline RL algorithms is an interesting area of future work.

\begin{table*}[t]
\caption{\textbf{OPAL Performance Using Reward Predicted After N Rounds of Querying}: Offline RL performance using rewards predicted with T-REX using a static number of randomly selected pairwise preferences (T-REX), Ensemble Disagreement(EnsemDis), Ensemble Information Gain (EnsemInfo), Dropout Disagreement (DropDis), Dropout Information Gain (DropInfo) and Ground Truth Reward (GT). N is set to 5 for Maze2D-Umaze, 10 for Maze2d-Medium and flow-merge-random, 15 for Hopper and Halfcheetah. N is selected based on the size of the dimension of observation space and the complexity of the environment. Results are averages over three random seeds. Results are normalized such that the ground truth performance is 100.0 and the random policy performance is 0.0. We report the performance of OPAL using AWR~\cite{peng2019awr} and CQL~\cite{kumar2020cql} for policy optimization.} 
\label{tab:finalreturn_awr_cql}
\vskip 0.15in
\begin{center}
\begin{small}
\begin{sc}
\begin{tabular}{l|rrrrr}
\toprule
AWR~\cite{peng2019awr}& & \multicolumn{4}{c}{Query Aquisition Method} \\
& T-REX &  EnsemDis & EnsemInfo & DropDis & DropInfo\\
\midrule
maze2d-umaze & 78.2 & \textbf{93.5} & 89.2 & 88.0 & 61.3 \\
maze2d-medium & 71.6 & \textbf{86.4} & 71.0 & 52.6 & 44.4 \\
hopper & 69.0 & 77.5 & 72.8 & 87.6  & \textbf{90.6}\\
halfcheetah & 96.1 & \textbf{113.7} & 100.6 & 84.4 & 91.7 \\
flow-merge-random & \textbf{110.1} & 89.0 & 92.1 & 86.2 & 84.2 \\
kitchen-complete & 79.6 & 105.0 & \textbf{158.8} & 48.5 & 65.4  \\
\bottomrule
\toprule
CQL \cite{kumar2020cql} & & \multicolumn{4}{c}{Query Aquisition Method} \\
& T-REX &  EnsemDis & EnsemInfo & DropDis & DropInfo\\
\midrule
maze2d-umaze & 87.0 & 54.8 & 100.4 & \textbf{102.3} & 95.2 \\
maze2d-medium & 41.1 & 33.1 & \textbf{115.4} & 1.6 & 34.5\\
hopper & 32.1 & 28.7 & 17.5 & 35.9  & \textbf{42.0}\\
halfcheetah & 68.4 & 75.9 & 76.3 & 75.0 & \textbf{79.6}\\
flow-merge-random & -13.6 & 44.3 & 103.2 & 69.6 & \textbf{114.0}\\
kitchen-complete & 85.7 & 100.0 & 71.4 & \textbf{114.3} & 71.4 \\
\bottomrule
\end{tabular}
\end{sc}
\end{small}
\end{center}
\vskip -0.1in
\end{table*}

\section{Franka Kitchen Learning Curves}
\begin{figure}[H]
     \centering
         \centering
        \includegraphics[width =0.35\textwidth]{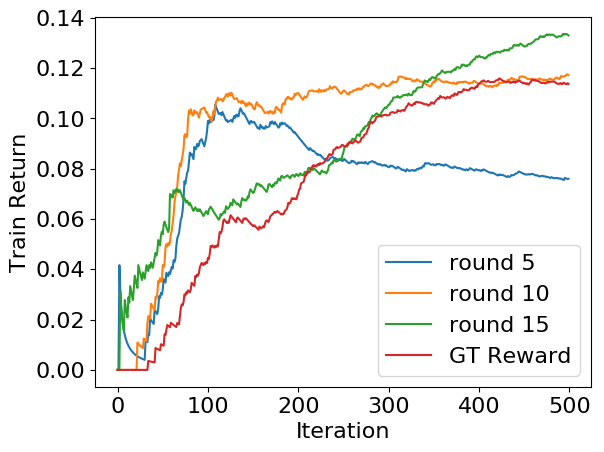}
         \caption{Kitchen-Complete-v1}
    \caption{Dropout disagreement after 15 rounds has similar performance to ground truth reward in Kitchen-complete}
     \label{fig:kitchen}
\end{figure}

The FrankaKitchen domain involves interacting with various objects in the kitchen to reach a certain state configuration. The objects you can interact with includes a water kettle, a light switch, a microwave, and cabinets. Our results in Table~\ref{tab:finalreturn_ablation} show that setting the reward to all zero or a constant reward causes the performance to degrade significantly.
However, the results in Table~\ref{tab:finalreturn} show that OPAL using Ensemble Disagreement and Ensemble InfoGain are able to shape the reward in a way that leads to better performance than using the ground-truth reward for the task. In Figure~\ref{fig:kitchen}, we plot the learning curves over time and find that OPAL also converges to a better performance faster. While surprising, these results support recent work showing that the shaping reward learned from pairwise preferences can boost the performance of RL, even when the agent has access to the ground truth reward function~\cite{memarian2021self}.

\section{Pairwise Preference Accuracy}
Pairwise preference accuracy on a held out set of trajectory paris for each approach is provided in Table~\ref{tab:rankacc}. Notably, when using randomly chosen queries (T-REX) the accuracy barely improves after round 5 in terms of ranking accuracy whereas all of our approaches continue to improve after round 5.

\section{Reward Model, Hyperparameter, And Dataset Details}
During active reward learning, a total of 7 models are used in the ensemble variant. For the reward models, we used  a  standard  MLP  architecture using hidden sizes [512, 256, 128, 64, 32] with  ReLU  activations.
For the dropout variant, we use 30 pass through of the model to calculate the posterior. Both the number of ensemble members and the number of pass-throughs were tuned to balance between speed and giving an accurate disagreement or information gain. 


For policy learning with AWR, lower dimensional environments including Maze2D-Umaze, Maze2D-Medium, and Hopper are ran with 400 iterations. Higher dimensional environments including Halfcheetah, Flow-Merge-Random, and Kitchen-Complete are ran with 1000 iterations. The number of iterations are tuned so that the policy converges before the maximum number of iterations. For CQL, policy learning rate is 1e-4, lagrange threshold is -1.0, min q weights is 5.0, min q version is 3, and policy eval start is 0. These numbers are suggested by the author of the CQL paper. 

All results are collected and averaged across 3 random seeds. All models are trained on an Azure Standard NC24 Promo instance, with 24 vCPUs, 224 GiB of RAM and 4 x K80 GPU (2 Physical Cards).
Datasets used to generate the customized behaviors (e.g. Constrained Goal Navigation, Orbit Policy, and Cartpole Windmill) will be released upon conference acceptance.

\begin{table*}[h]
\caption{Pairwise preference accuracy on held out set of trajectory pairs after 5 rounds and 10 rounds of queries respectively for different types of query acquisition functions.}
\label{tab:rankacc}
\vskip 0.15in
\begin{center}
\begin{small}
\begin{sc}
\begin{tabular}{lccccr}
\toprule
& Random &  EnsemDis & EnsemInfo & DropDis & DropInfo\\
\midrule
maze2d-umaze & 0.818, 0.867 & 0.817, 0.916 & 0.77,  0.862 & 0.874,  0.893 & 0.715,  0.755 \\
maze2d-medium & 0.646, 0.686 & 0.650, 0.783 & 0.638,  0.824 & 0.636,  0.795 & 0.692,  0.756 \\
hopper & 0.929, 0.922 & 0.943, 0.966 & 0.952,  0.966 & 0.946,  0.966  & 0.927,  0.960 \\
halfcheetah & 0.814, 0.873 & 0.852, 0.942 & 0.866,  0.944 & 0.875,   0.927 & 0.833,  0.913 \\
flow-merge-random & 0.808, 0.841 & 0.829, 0.881 & 0.825,  0.861 & 0.846,  0.884 & 0.839,  0.893 \\
kitchen & 0.964, 0.978 & 0.981, 0.986 & 0.967, 0.984 & 0.973, 0.990 & 0.953, 0.986 \\
\bottomrule
\end{tabular}
\end{sc}
\end{small}
\end{center}
\vskip -0.1in
\end{table*}

\section{Pseudocode}

\begin{algorithm}[H]
\caption{OPAL}
\label{alg:opal}
\begin{algorithmic}[1]
   \STATE {\bfseries Require:} A dataset $\mathcal{D} = \{\tau_0, ..., \tau_N\}$, where each trajectory $\tau_i$ consists of 
   $\tau_i $ = $\{(s_{i,0}, a_{i,0}, s'_{i,0})$, ..., $(s_{i,M}, a_{i,M}, s'_{i,M})\}$. \label{lst:line:blah1}
   \STATE // REWARD LEARNING \label{lst:line:blah2}
   \STATE Generate dataset of pairs of snippets $\mathcal{D}_{snip}$ 
  \STATE Initialize reward model $\hat{r}_{\psi}$ 
  \FOR{each iteration}
   \STATE Compute estimate of VOI across all pairs in $\mathcal{D}_{snip}$
   \STATE Find snippet pair ($\tau_{s,1}$, $\tau_{s,2}$) with highest
   estimated VOI
  \STATE Query preference label $y$
   \STATE Store query pair and label $\mathcal{D}_{rew} \leftarrow (\tau_{s,1}, \tau_{s,2}, y)$
   \STATE Train reward model with updated $\mathcal{D}_{rew}$
  \ENDFOR
   \STATE Label transitions in $\mathcal{D}$ with $\hat{r}_{\psi}$
   \STATE // POLICY LEARNING \label{lst:line:blah3}
   \STATE Run selected offline RL algorithm on $\mathcal{D}$
\end{algorithmic}
\end{algorithm}

\section{Table 1 Extended}
\begin{table*}[t]
\caption{Offline RL performance on D4RL benchmark tasks comparing the performance with the true reward to performance using a constant reward equal to the average reward over the dataset (Avg) and zero rewards everywhere (Zero). Even when all the rewards are set to the average or to zero, many offline RL algorithms still perform surprisingly well. In this table, we present the experiments ran with BCQ. The degradation percentage is calculated as ${\max(\text{GT} - \max(\text{AVG}, \text{ZERO}, \text{RANDOM}), 0)\over |\text{GT}| }\times 100\%$.
}
\label{tab:finalreturn_ablation_bcq}
\vskip 0.15in
\begin{center}
\begin{small}
\begin{sc}
\begin{tabular}{lrrrrrr}
\toprule
Task &  
GT &  Avg & Zero & Random & Degradation \% \\
\midrule
    flow-ring-random-v1 & 14.3 & -41.8 & -67.5 & -166.2 & 392.3\\
    flow-merge-random-v1 & 334 & 130.3 & 121.4 & 117 & 61.0 \\
    \midrule
    maze2d-umaze & 104.2 & 36.5 & 41.8 & 49.5 & 52.5 \\
    maze2d-medium & 106.6 & 23.3 & 28.5 & 44.8 & 58.0 \\
    \midrule
    halfcheetah-random & -0.6 & -1.5 & -1.9 & -285.8 & 133.9 \\
    halfcheetah-medium-replay & 3400.0 & 2659.3 & 3185.0 & -285.8 & 6.3\\
    halfcheetah-medium & 3057.7 & 2219.1 & 2651.7 & -285.8 & 13.3 \\
    halfcheetah-medium-expert & 10905.8 & 11146.9 & 11030.3 & -285.8 &  0.0\\
    halfcheetah-expert & 63.6 & 115.8 & -32.4 & -285.8 & 0.0\\
    \midrule
    hopper-random & 199.8 & 186.0 & 199.4 & 18.3 & 0.2 \\
    hopper-medium-replay & 1188.1 & 1082.4 & 908.3 & 18.3 & 8.9 \\
    hopper-medium & 922.3 & 745.4 & 842.2 & 18.3 & 8.7  \\
    hopper-medium-expert & 343.3 & 225.7 & 202.0 & 18.3 & 34.3 \\
    hopper-expert & 241.5 & 268.1 & 343.3 & 18.3 & 0.0 \\
    \midrule
    kitchen-complete & 0.07 & 0.11 & 0.04 & 0.0 & 0.0 \\
    kitchen-mixed & 0.08 & 0.07 & 0.02 & 0.0 & 12.5 \\
    kitchen-partial & 0.16 & 0.17 & 0.13 & 0.0 & 0.0 \\
\bottomrule
\end{tabular}
\end{sc}
\end{small}
\end{center}
\vskip -0.1in
\end{table*}

\begin{table*}[t]
\caption{Offline RL performance on D4RL benchmark tasks comparing the performance with the true reward to performance using a constant reward equal to the average reward over the dataset (Avg) and zero rewards everywhere (Zero). Even when all the rewards are set to the average or to zero, many offline RL algorithms still perform surprisingly well. In this table, we present the experiments ran with BEAR. The degradation percentage is calculated as ${\max(\text{GT} - \max(\text{AVG}, \text{ZERO}, \text{RANDOM}), 0)\over |\text{GT}| }\times 100\%$
}
\label{tab:finalreturn_ablation_bear}
\vskip 0.15in
\begin{center}
\begin{small}
\begin{sc}
\begin{tabular}{lrrrrrr}
\toprule
Task &  
GT &  Avg & Zero & Random & Degradation \% \\
\midrule
    flow-ring-random-v1 & 13.4 & 18.2 & 11.6 & -166.2 & 0.0\\
    flow-merge-random-v1 & 75.9 & 62.1 & 68.3 & 117.0 & 0.0 \\
    \midrule
    maze2d-umaze & 66.6 & 44.8 & 59.7 & 49.5 & 10.3 \\
    maze2d-medium & 18.6 & 29.8 & 22.5 & 44.8 & 0.0 \\
    \midrule
    halfcheetah-random & 0.8 & -0.9 & -0.5 & -285.8 & 160.5 \\
    halfcheetah-medium-replay & 4997.5 & 3919.6 & 4282.7 & -285.8 & 14.3\\
    halfcheetah-medium & 5260.9 & 4910.3 & 4941.4 & -285.8 & 6.1 \\
    halfcheetah-medium-expert & 5346.9 & 5033.4 & 4709.1 & -285.8 &  5.9\\
    halfcheetah-expert & 11329.1 & 11268.9 & 10627.6 & -285.8 & 0.5\\
    \midrule
    hopper-random & 221.9 & 215.6 & 86.3 & 18.3 & 2.9 \\
    hopper-medium-replay & 1261.9 & 659.0 & 1311.2 & 18.3 & 0.0 \\
    hopper-medium & 1564.3 & 1749.3 & 1601.5 & 18.3 & 0.0 \\
    hopper-medium-expert & 2269.7 & 2995.2 & 1387.9 & 18.3 & 0.0 \\
    hopper-expert & 2110.6 & 2925.9 & 2826.3 & 18.3 & 0.0 \\
    \midrule
    kitchen-complete & 0.8 & 0.7 & 0.1 & 0.0 & 12.0 \\
    kitchen-mixed & 0.7 & 0.6 & 0.2 & 0.0 & 14.3 \\
    kitchen-partial & 0.2 & 0.3 & 0.1 & 0 & 0.0 \\
\bottomrule
\end{tabular}
\end{sc}
\end{small}
\end{center}
\vskip -0.1in
\end{table*}

\begin{table*}[t]
\caption{Offline RL performance on D4RL benchmark tasks comparing the performance with the true reward to performance using a constant reward equal to the average reward over the dataset (Avg) and zero rewards everywhere (Zero). Even when all the rewards are set to the average or to zero, many offline RL algorithms still perform surprisingly well. In this table, we present the experiments ran with CQL. The degradation percentage is calculated as ${\max(\text{GT} - \max(\text{AVG}, \text{ZERO}, \text{RANDOM}), 0)\over |\text{GT}| }\times 100\%$.
}
\label{tab:finalreturn_ablation_cql}
\vskip 0.15in
\begin{center}
\begin{small}
\begin{sc}
\begin{tabular}{lrrrrrr}
\toprule
Task &  
GT &  Avg & Zero & Random & Degradation \% \\
\midrule
    flow-ring-random-v1 & 13.4	& -49.5	& -23.3 & -166.2 & 274.1\\
    flow-merge-random-v1 & 156.1 & 98.1 & 50.5 & 117.0 & 25.0\\
    \midrule
    maze2d-umaze & 83.4 & 46.0 & 38.2 & 49.5 & 40.7 \\
    maze2d-medium & 107.9 & 34.3 & 19.6 & 44.8 & 58.5 \\
    \midrule
    halfcheetah-random & 3140.7 & -281.5 & -158.5 & -285.8 & 105.0 \\
    halfcheetah-medium-replay & 5602.9 & 1145.4 & -512.2 & -285.8 & 79.6\\
    halfcheetah-medium & 5546.9 & 5130.2 & 5037.2 & -285.8 & 7.5 \\
    halfcheetah-medium-expert & 3544.6 & 4372.7 & 5505.3 & -285.8 &  0.0 \\
    halfcheetah-expert & 258.2 & 45.5 & 3546.6 & -285.8 & 0.0 \\
    \midrule
    hopper-random & 968.9 & 512.0 & 11.0 & 18.3 & 47.2 \\
    hopper-medium-replay & 2484.2 & 1970.3 & 563.7 & 18.3 & 20.7 \\
    hopper-medium & 1709.9 & 2205.4 & 2098.7 & 18.3 & 0.0 \\
    hopper-medium-expert & 1156	& 1136.9 & 1210.9 & 18.3 & 0.0 \\
    hopper-expert & 919.1 & 904.1 & 2092.7 & 18.3 & 0.0 \\
    \midrule
    kitchen-complete & 0.8 & 0.7 & 0.6 & 0.0 & 14.3 \\
    kitchen-mixed & 0.6 & 0.4 & 0.8 & 0.0 & 0.0 \\
    kitchen-partial & 0.3 & 0.7 & 0.4 & 0.0 & 0.0 \\
\bottomrule
\end{tabular}
\end{sc}
\end{small}
\end{center}
\vskip -0.1in
\end{table*}

\end{document}